\documentclass[sigconf]{acmart}

\usepackage{multirow}
\usepackage{hyperref}
\usepackage{siunitx}
\usepackage{float}
\usepackage{subcaption}
\usepackage{enumitem}

\setlist[itemize]{leftmargin=*}

\newenvironment{notalistitem}
{\begin{itemize}
 
\item}
{\end{itemize}}
\AtBeginDocument{%
  \providecommand\BibTeX{{%
    \normalfont B\kern-0.5em{\scshape i\kern-0.25em b}\kern-0.8em\TeX}}}

\copyrightyear{2023} 
\acmYear{2023} 
\setcopyright{acmlicensed}\acmConference[ICAIF '23]{4th ACM International Conference on AI in Finance}{November 27--29, 2023}{Brooklyn, NY, USA}
\acmBooktitle{4th ACM International Conference on AI in Finance (ICAIF '23), November 27--29, 2023, Brooklyn, NY, USA}
\acmPrice{15.00}
\acmDOI{10.1145/3604237.3626908}
\acmISBN{979-8-4007-0240-2/23/11}

\begin{document}
% \tableofcontents
\title{FlowMind: Automatic Workflow Generation with LLMs}

\author{Zhen Zeng}
\affiliation{%
  \institution{J.~P.~Morgan AI Research}
  \city{New York}
  \state{NY}
  \country{USA}}
\email{zhen.zeng@jpmchase.com}

\author{William Watson}
\affiliation{%
  \institution{J.~P.~Morgan AI Research}
  \city{New York}
  \state{NY}
  \country{USA}}
\email{william.watson@jpmchase.com}

\author{Nicole Cho}
\affiliation{%
  \institution{J.~P.~Morgan AI Research}
  \city{New York}
  \state{NY}
  \country{USA}}
\email{nicole.cho@jpmorgan.com}

\author{Saba Rahimi}
\affiliation{%
  \institution{J.~P.~Morgan AI Research}
  \city{New York}
  \state{NY}
  \country{USA}}
\email{saba.rahimi@jpmorgan.com}

\author{Shayleen Reynolds}
\affiliation{%
  \institution{J.~P.~Morgan AI Research}
  \city{New York}
  \state{NY}
  \country{USA}}
\email{shayleen.reynolds@jpmchase.com}

\author{Tucker Balch}
\affiliation{%
  \institution{J.~P.~Morgan AI Research}
  \city{New York}
  \state{NY}
  \country{USA}}
\email{tucker.balch@jpmchase.com}

\author{Manuela Veloso}
\affiliation{%
  \institution{J.~P.~Morgan AI Research}
  \city{New York}
  \state{NY}
  \country{USA}}
\email{manuela.veloso@jpmchase.com}
\renewcommand{\shortauthors}{Zeng, et al.}





% \begin{figure}
%   \includegraphics[width=\textwidth]{samples/sampleteaser}
% \end{figure}


\begin{abstract}
The rapidly evolving field of Robotic Process Automation (RPA) has made significant strides in automating repetitive processes, yet its effectiveness diminishes in scenarios requiring spontaneous or unpredictable tasks demanded by users. This paper introduces a novel approach, \textbf{FlowMind}, leveraging the capabilities of Large Language Models (LLMs) such as Generative Pretrained Transformer (GPT), to address this limitation and create an automatic workflow generation system. In FlowMind, we propose a generic prompt recipe for a lecture that helps ground LLM reasoning with reliable Application Programming Interfaces (APIs). With this, FlowMind not only mitigates the common issue of hallucinations in LLMs, but also eliminates direct interaction between LLMs and proprietary data or code, thus ensuring the integrity and confidentiality of information — a cornerstone in financial services. FlowMind further simplifies user interaction by presenting high-level descriptions of auto-generated workflows, enabling users to inspect and provide feedback effectively. We also introduce \textbf{NCEN-QA}, a new dataset in finance for benchmarking question-answering tasks from N-CEN reports on funds. We used NCEN-QA to evaluate the performance of workflows generated by FlowMind against baseline and ablation variants of FlowMind. We demonstrate the success of FlowMind, the importance of each component in the proposed lecture recipe, and the effectiveness of user interaction and feedback in FlowMind.
\end{abstract}

\maketitle

\begin{figure}[t!]
  \centering
  % \includegraphics[width=0.45\textwidth]{example-image-a}
  % Answer: [trim={left bottom right top},clip]
  \includegraphics[clip, trim=9.2cm 5.9cm 10.5cm 4.4cm, width=0.47\textwidth]{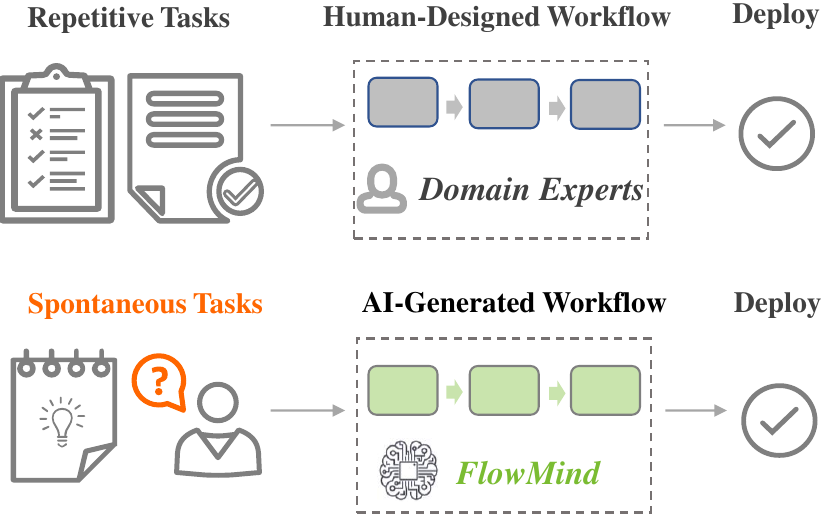}
  \caption{FlowMind automates spontaneous tasks demanded by users through on-the-fly workflow generation, advancing beyond traditional automation of repetitive tasks designed by domain experts.}
  \label{fig:teaser}
\end{figure}

\section{Introduction}
% key contributions:
% - we provide a generic recipe for the prompt template for workflow generation, and demonstrated the importance of each component through an ablation study \sre{do we want to  speak in present or past tense? I.e. "we provided", "and demonsrated" v. "we provide"..etc.} -- check if Langchain provides that template, HuggingFace does have the template in their paper, but no public evaluation on the importance of each component
% - we don't assume the type of functions/APIs to be used (huggingface is all transformers, langchain is not limited but?)
% - we benchmarked the performance on financial datasets
% - we innovate on on-loop feedbacks from user before executing

The paradigm of Robotic Process Automation (RPA) has vastly transformed the landscape of task execution by enabling the automation of repetitive processes. However, its reliance on expert knowledge and well-defined procedures can fall short in situations where more spontaneous or unpredictable tasks arise. To address these challenges, we explore the capabilities of Large Language Models (LLMs) and present a framework that leverages their potential to create a versatile, dynamic, and secure workflow generation system.

In this paper, we introduce a novel approach, \textbf{FlowMind}, that enables automatic workflow generation using LLMs, following the proposed generic lecture recipe of prompt design. Through an extensive and rigorous study, we dissect each component of our prompt design to demonstrate its importance and contributions to the overall effectiveness of automatic workflow generation. FlowMind allows us to harness the vast capabilities of LLMs, specifically Generative Pretrained Transformer (GPT) model, in a more defined and structured manner, leading to robust and efficient code generation for workflow execution.

A key feature of our framework lies in its robustness against hallucinations often experienced with LLMs. We ground the reasoning of LLMs with the aid of Application Programming Interfaces (APIs). These APIs are reliable functions developed and tested by domain experts, ensuring their accuracy and reliability. Proprietary software developed in industry is typically composed of such reliable APIs. FlowMind is able to leverage APIs provided to it while ensuring that the LLMs do not directly interact with any proprietary code or data, protecting code and data privacy. This protection is achieved by allowing LLMs to act only on the high-level descriptions of the APIs, enhancing security and ensuring a reliable generation of workflows.

Understanding the necessity for human oversight, our system also integrates user feedback. Without assuming the programming experiences of the user, the system provides a high-level description of the auto-generated workflow, allowing novice users to inspect and provide feedback. FlowMind then takes the user feedback and adjusts the generated workflow if needed. This two-way interaction empowers users to enhance the workflow based on their knowledge and the unique demands of their tasks, thus enhancing the flexibility and adaptability of our system.

As part of our contribution to the broader research community, we provide a benchmark dataset \textbf{NCEN-QA} in finance, built on top of N-CEN reports\footnote{\url{https://www.sec.gov/files/formn-cen.pdf}}. 
This dataset serves as a robust platform for evaluating workflow generation systems, particularly in question-answering tasks in the domain of funds. Our experiments on \textbf{NCEN-QA} demonstrate the effectiveness of our proposed method. We believe this dataset will open new avenues for performance comparisons and drive the advancement of workflow generation research in finance.

Overall, our contributions are three-fold:
\begin{itemize}
    \item We propose \textbf{FlowMind}, a novel approach that facilitates automatic workflow generation using LLMs, which addresses hallucination and data privacy concerns, underpinned by a generic lecture recipe of prompt design.
    \item Our system incorporates a feedback mechanism enabling users to effectively inspect and provide inputs to refine the generated workflow when needed.
    \item We offer a benchmark dataset \textbf{NCEN-QA} in the field of Finance for evaluating workflow generation in handling question-answering tasks on funds.
\end{itemize}

Next, we will introduce and discuss the relevant literature in section~\ref{sec:related_work}, the proposed FlowMind method~\ref{sec:method}, as well as experiments results and analysis in section~\ref{sec:experiments}.

\begin{figure*}[t!]
  \centering
  % \includegraphics[width=0.8\textwidth]{figures/API_powered_LLM_single_diagram.png}
  % Answer: [trim={left bottom right top},clip]
  \includegraphics[clip, trim=6.55cm 7.4cm 6.55cm 6.55cm, width=0.8\textwidth]{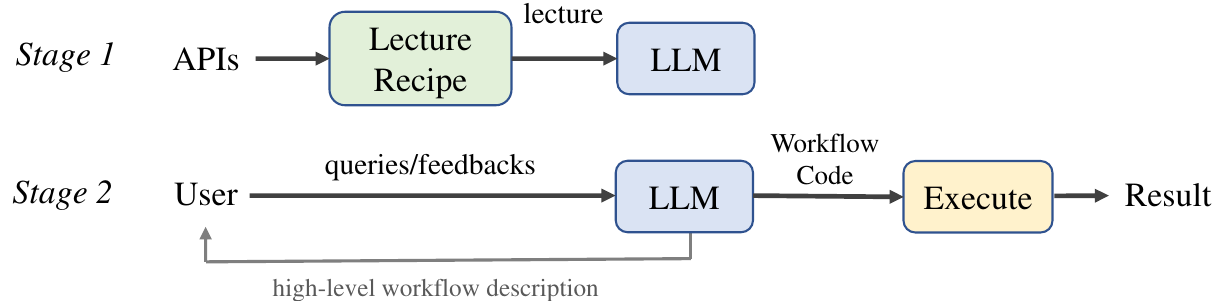}
  \caption{Overview of FlowMind framework. (a) Stage 1: we follow the proposed generic \textit{lecture} recipe to generate a lecture prompt, which educates the LLM about the context, APIs, and get ready to write code; (b) Stage 2: LLM can then take user queries or tasks and auto-generate the workflow code that makes use of the introduced APIs. The workflow code is executed to deliver the result. During stage 2, we enable a feedback loop between FlowMind and the user, where FlowMind provides high-level description of the generated workflow in plain-language, and the user inputs feedback to FlowMind to approve or refine the workflow if needed.}
  \label{fig:overview}
\end{figure*}

\section{Related Work}\label{sec:related_work}
% we can use https://www.connectedpapers.com/ to explore connected papers

\subsection{Robotic Process Automation}
% 1. Robotic Process Automation (RFP): in most RFP works, the targets are repetitive processes (give examples) in industry. For these tasks, it is effective to pre-define the workflow to be reused for process automation [cite]. However, it requires expert knowledge and well-defined procedures or steps in the process. For spontaneous tasks/requests that are demanded by users interactively, we cannot prescribe the corresponding workflow beforehand, instead, we need automatic workflow generation to effectively handle them.

Robotic Process Automation (RPA) has been widely recognized for its potential to automate repetitive tasks in various industries~\cite{10.1145/3275116.3275129, 8754139, villar2021robotic}. For instance, repetitive tasks such as data entry, invoice processing, or customer service responses have been successfully automated using RPA~\cite{vanderAalst2018, Hofmann2020}. These tasks often involve pre-defined workflows that, once created, can be reused multiple times, significantly enhancing the efficiency and accuracy of the tasks. Based on rule-based business processes, or by observing human digital actions, RPAs provide a software solution to automating repetitive tasks and has seen success across various domains. However, this approach requires expert knowledge and well-defined procedures or steps~\cite{10.1007/978-3-030-58666-9_27}, which are not always available, especially for more spontaneous or unpredictable tasks. Additionally, RPAs lack objective reasoning around its application or development~\cite{SYED2020103162}. When dealing with such tasks, which users might interactively demand, the limitation of RPA becomes apparent. Pre-defined workflows are inadequate for handling such situations, necessitating automatic workflow generation to manage them effectively.

\subsection{LLMs for Coding}
% 2. LLMs for Coding: Workflow can be described and executed in the form of code, such code written in programming language (e.g. python). Existing research have explored using LLMs for code generation [cite]. More recently, with the rise of GPT [cite], researchers have explored particularly the effectiveness of GPT auto-generating code in different domains including Robotics [cite ChatGPT for Robotics, Code as Policies from Google], and so on. These works have shed lights on promising ways such as prompt engineering and few shots learning to leverage the advances of GPT for code generation. However, most of recent studies provided only qualitative examples/demos, there wasn't rigorous analysis over quantitative performance on benchmark datasets, especially in Finance [check]. We are the first to provide such benchmark dataset in Finance, as well as the performance of our proposed method, its variants, and competing methods with clear metrics.

Language Models (LLMs), especially in their use for code generation, have seen considerable exploration and advancement~\cite{nijkamp2022codegen, vaithilingam2022expectation, poesia2022synchromesh}. The rise of the Generative Pretrained Transformer (GPT) models~\cite{radford2018improving, radford2019language} has spurred further exploration, specifically regarding their potential to generate code in various domains~\cite{liu2023your, vemprala2023chatgpt}. Furthermore, prior work has explored chain of thought through code, as demonstrated in~\cite{liang2023code} for robotic programs, action plan generation~\cite{saycan2022arxiv}, web browsing~\cite{nakano2022webgpt}, learning tools~\cite{Schick2023ToolformerLM}, or generating valid arithmetic programs~\cite{Gao2022PALPL}. Recently, LLMs have shown the ability to construct modular code for visual question answering based on abstractions of high-level APIs~\cite{subramanian2023modular}. These studies utilized techniques like prompt engineering and few-shot learning to harness the capabilities of GPT for code generation. However, while promising, much of this work has been somewhat anecdotal, providing qualitative examples and demonstrations but lacking rigorous analysis of quantitative performance against benchmark datasets. Few LLMs have been fine-tuned on finance-related tasks~\cite{wu2023bloomberggpt, zhang2023instruct, yang2023fingpt, liu2023fingpt}, however, none of these focused on leveraging LLMs for coding. Our work leverages LLMs for coding workflow structures and we introduce a new dataset in Finance for rigorous evaluation.

\subsection{Workflow Generation with LLMs}
% 3. Application Developments with LLMs: Building on LLMs' ability of code generation, there is a recent emergenc   e of application development tools built on LLMs that are suitable for workflow generation, including Langchain [cite], HuggingFace Transformer Agent [cite], AutoGPT [cite], and so on. Langchain focuses on combining LLMs with external data, where they use pinecone to split data into chunks for scalability. Their system allows LLMs to interact with a large database. However, data privacy presents a big challenge, especially in the Finance industry. In general, we should limit the direct interaction between LLMs and private data as much as possible [reason], especially for advanced LLMs like GPT where the inference doesn't happen locally. Transformer Agent is targeted at automatically combining different transformer models for user-requested tasks, thus not necessarily flexible enough to easily incorporate other types of functions/models developed as proprietary software in the industry. In contrast, we provide a generic recipe for workflow generation that enables GPT to handle user requests without data privacy intrusion or assumptions on the types of functions/models to combine during code generation.

Building on the capabilities of LLMs for code generation, recent advancements have seen the development of application tools that leverage LLMs for workflow generation, such as Langchain~\cite{langchain}, HuggingFace's Transformer Agent~\cite{transAgent}, and AutoGPT~\cite{AutoGPT}. Langchain, for instance, focuses on integrating LLMs with external data, utilizing a scalable approach to chunk data for interaction with a large database. However, applying this in fields like finance raises serious data privacy concerns because the model directly interacts with the data. With powerful LLMs like GPT, where inference doesn't occur locally, it is critical to limit direct interaction with private data as much as possible. On the other hand, Transformer Agent, while being a novel approach to task automation, is primarily aimed at combining different transformer models for user-requested tasks. Thus, it might not be flexible enough to easily incorporate other types of functions/models typically developed as proprietary software in various industries. AutoGPT provides limited problem-solving capabilities due to the limited set of functions in its library, such as web browsing and executing code. Our proposed method, in contrast to the aforementioned tools, offers a more generic and adaptable approach to workflow generation, allowing user requests to be handled without 1) breaching data privacy or 2) making assumptions about the types of functions/models to be combined during code generation.

% Our work aims to fill this gap by being the first to provide a benchmark dataset in Finance that can effectively evaluate workflow generation on question-anwering tasks, alongside a rigorous evaluation of our proposed method on automatic workflow generation, its ablation variants, and competing method.

\section{Method}\label{sec:method}
The FlowMind framework functions in two primary stages as illustrated in Figure~\ref{fig:overview}: 1) initial lecture to the LLM on the task context and available APIs, and 2) subsequent workflow generation using the APIs and workflow execution to deliver the result to user. During stage 2, we enable an optinal feedback loop between FlowMind and the user such that user can easily review and provide feedback to FlowMind, and FlowMind can adjust the generated workflow accordingly.

\subsection{Lecture to LLM}\label{sec:recipe}
\begin{figure*}[t!]
  \centering
  % \includegraphics[width=0.9\textwidth, height=0.35\textwidth]{example-image-b}
  % Answer: [trim={left bottom right top},clip]
  \includegraphics[clip, trim=1.75cm 2.95cm 2.35cm 4.0cm, width=0.9\textwidth]{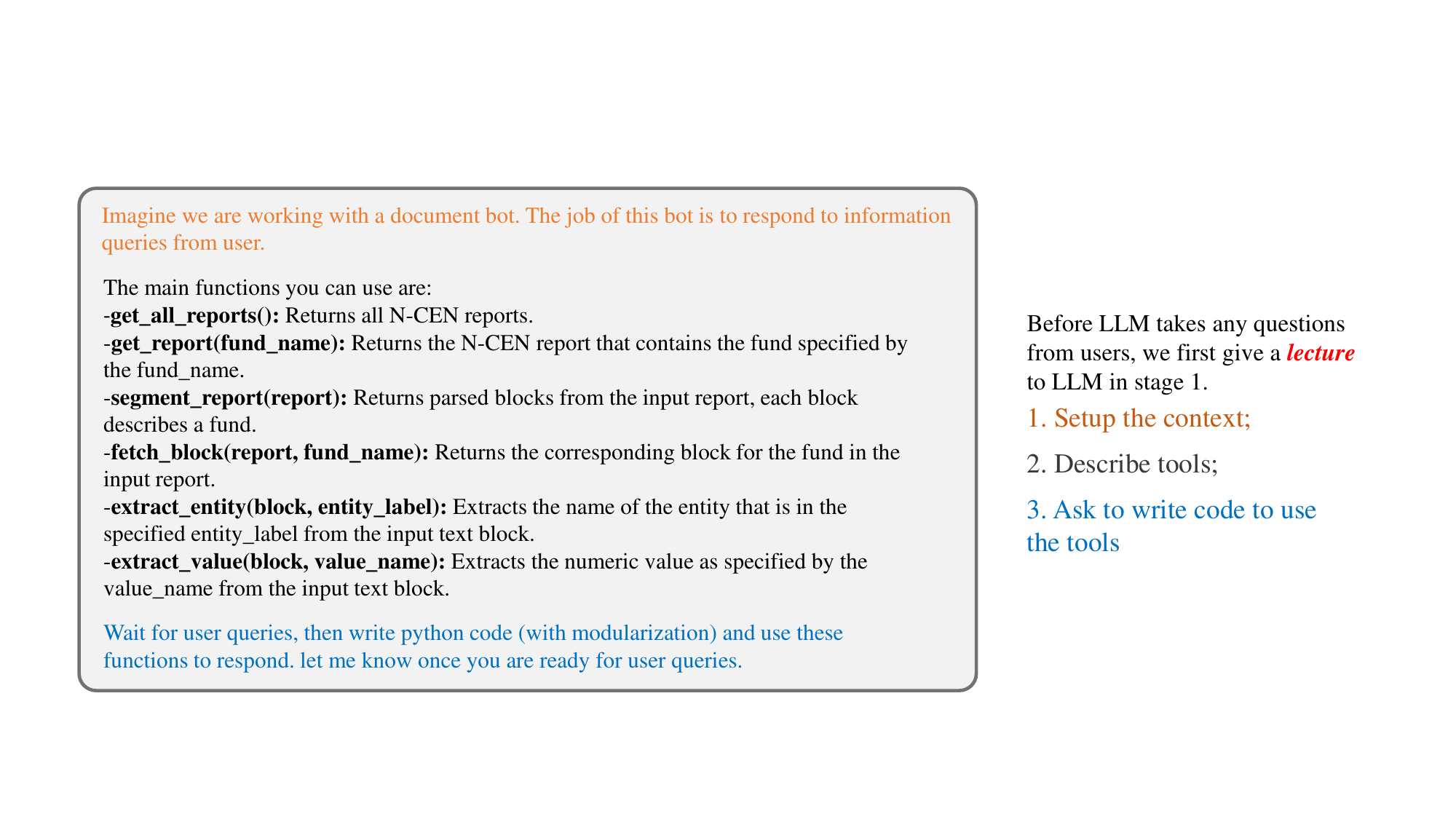}
  \caption{Before an LLM takes any queries or tasks from users, we first give a \textit{lecture} to it. We show an example of such a lecture above. The proposed generic lecture recipe includes: 1) setting up the context, 2) enumerating the available APIs with each function declaration, parameters, and high-level descriptions, and 3) prompting the LLM to write workflow code using these APIs.}
  \label{fig:recipe}
\end{figure*}

The first stage of the FlowMind framework involves a \textit{lecture} on the context, available APIs, and the need to generate workflow code for the LLM. We adhere to our proposed \textbf{generic lecture recipe} to generate an informative lecture on the context and APIs, ensuring the LLM has a clear understanding of the overall goal, as well as the scope, inputs, and outputs of the functions in the APIs. The lecture recipe is crafted with three components, each with a distinct role. Specifically, the proposed lecture prompt recipe covers:
\begin{itemize}
    \item \textit{Context}: First we introduce the context which covers the domain of the expected tasks/queries from the user. For example, in our experiments, we set up the context as handling information queries from user, as shown in Figure~\ref{fig:recipe}.
    \item \textit{APIs}: Then we provide a list of structured descriptions of the available APIs to use for the LLM. Importantly, we introduce the name of the function, the input arguments, and the output variables. Note that the function names, input arguments, and output descriptions must be semantically meaningful and relevant to the context above such that the LLM can comprehend to make good use of the functions.
    \item \textit{Code}: Lastly we ask the LLM to prepare to write the workflow code using the provided APIs upon receiving user query/task.
\end{itemize}
An example of such a lecture in our experiments is shown in Figure~\ref{fig:recipe}. The crafted prompt following the lecture recipe enables the LLM to gain the necessary understanding of the context and available APIs, to utilize them in the subsequent stage of workflow generation effectively. We highlight the effect of each component in the lecture recipe in our experiments.

\subsection{Workflow Generation and Execution}
In the second stage, LLM leverages the API knowledge gained from the first stage to take user queries or tasks and generate corresponding workflow code. This stage involves two key components: code generation and code execution. In code generation, LLM creates a workflow, making use of the introduced APIs to address the user's query or task effectively. The workflow is then executed to generate the output to user. We show examples of the auto-generated workflows by FlowMind and derived answers to several sample user questions in our Experiments (Figure~\ref{fig:easy},~\ref{fig:intermediate},~\ref{fig:hard}).

A distinct feature of FlowMind is the ability to take user feedback during the second stage. The system presents a high-level description of the generated workflow to the user, enabling users to understand the workflow's functionality and structure without the need to closely examine the underlying code. This allows the users to effectively provide feedback on the generated workflow, which the LLM can then incorporate to refine the workflow if necessary, ensuring that the system accurately addresses the user's needs. We prompt LLM with "Could you provide a concise high-level summary of the flow of code? Then take feedback to see if code needs to be updated" to get the high-level workflow description. Examples are discussed and shown in Experiments (Figure~\ref{fig:feedback}).

% In conclusion, the FlowMind framework provides a structured, iterative approach to automatic workflow generation, ensuring the efficient and effective use of LLMs. By allowing for ongoing user feedback and maintaining a strong focus on data security and code reliability, FlowMind sets a high standard in automatic workflow generation, paving the way for significant advancements in this dynamic field.

\section{Experiments}\label{sec:experiments}
To evaluate the effectiveness of FlowMind, we propose a benchmark dataset, NCEN-QA, derived from N-CEN reports on funds in the domain of Finance. In addition, N-CEN APIs were provided to FlowMind to consume and extract information from N-CEN reports. We used GPT as the LLM in the proposed framework.

By lecturing FlowMind on the N-CEN APIs, we carried out rigorous assessments of FlowMind's ability to auto-generate workflows to handle question-answering tasks in NCEN-QA. We demonstrate that FlowMind, even without user feedbacks, already significantly outperformed GPT question answering based on context retrieval, a commonly adopted methodology in the literature~\cite{rubin2021learning, ram2023context, pereira2023visconde}. Our ablation studies reveal the importance of each component in our lecture prompt recipe introduced in Section~\ref{sec:recipe}. Furthermore, we show that FlowMind's performance can be further boosted when user feedbacks are incorporated.

Next, we first introduce N-CEN reports in the finance domain and how we composed the NCEN-QA dataset in Section~\ref{sec:ncen-qa}, discuss the N-CEN APIs in Section~\ref{sec:apis}, then the benchmarked methods in Section~\ref{sec:benchmarks}, and experiments results in Section~\ref{sec:results}. Throughout the experiments, we used \texttt{gpt-3.5-turbo} with temperature 0 to minimize randomness.

\subsection{NCEN-QA}\label{sec:ncen-qa}
% background on N-CEN
We composed the NCEN-QA dataset by creating $600$ question-answer pairs centered on fund data found in N-CEN reports. N-CEN reports are mandatory annual filings for registered investment companies in the US, with each report submitted on a trust level, meaning each company could submit one report representing numerous funds. These reports provide a wealth of information on a range of funds, including their custodians, pricing services, investment advisors, gross commission, fund net assets, etc.

We crawled the most recently filed N-CEN reports from each reporting investment company on the public database Edgar~\cite{edgar} from the U.S. Securities and Exchange Commission (SEC) site. We scraped the reports across the past 3 years (prior to May 11th, 2023), at a maximum throughput of 10 reports per sec (SEC site limits for a single host), yielding 8,548 reports. Among these reports, there are duplicates of reported funds, and we cleaned the data to retain the most recent filing of each fund. As a result, we gathered 2,794 reports which covered a grand total of 12k funds. Based on the crawled data, we then composed sets of fund questions at varying difficulty levels \textbf{NCEN-QA-Easy}, \textbf{NCEN-QA-Intermediate}, and \textbf{NCEN-QA-Hard}, each containing $200$ question-answer pairs, as explained below.

\subsubsection{\textbf{NCEN-QA-Easy}} 

% \sra{make all the sample quesitons in the paper left aligned instead of center aligned; make it like "question template:    ", then " sample Q:   ", "sample A:    "; nd make these italic; you can also pick some color for each, so it's easier to see the pattern of these sample question blocks across the paper, like dark orange for "question template",  then orange for "sample Q: ", then blue for "sample A: "}

In this set, each question focuses on a single piece of information about a fund, requiring the investigation of only one N-CEN report in which the fund was reported, and in particular the specific block of texts describing the queried fund.
%one N-CEN report and the extraction of a single-entry answer. %The simplicity of these questions lies in their directness, where the answer includes the name of a single entity.
Example questions include:
\begin{notalistitem}
Q1: Who is the custodian for the Precious Metals Mutual Fund?
\end{notalistitem}
\begin{notalistitem}
Q2: What is the gross commission for the Rule One Fund?
\end{notalistitem}

% simple question template 1: "Who are the {custodian} for fund {X}?"
% sample Q.1.: "Who are the custodians for the PRECIOUS METALS FUND?"\\
% sample A.1.: "U.S. BANK NATIONAL ASSOCIATION"\\ 
% simple question template 2: "What was the {gross commission} for fund {X}?"\\
% sample Q.2.: "What was the gross commission for the WCM SMALL CAP GROWTH FUND?"\\
% sample A.2.: "20338.0"   

We sampled $200$ funds and derived the answers accordingly from N-CEN reports, encompassing both types of questions as shown in Q1 and Q2. Q1-type questions are focused on entities that provide certain services for the queried fund, including the custodian, investment advisor, collateral manager, administrator, and pricing services. On the other hand, Q2-type questions are focused on numbers such as gross commission and purchase sales. We kept the 200 question-answer pairs roughly evenly spread over the aforementioned fund information.

\subsubsection{\textbf{NCEN-QA-Intermediate}} 
In this set, each question also focuses only on one fund. However, the question is slightly more complex than the ones in NCEN-QA-Easy because it requires mathematical operations (e.g. division) on top of multiple pieces of information regarding the queried fund.
%involve exploring a mathematical relationship over two different relations for a single fund. Like Simple NCEN-QA, this subset also centers on investigating one specific fund, and the answers consist of a single value. These questions require a bit more \sre{can probably remove "a bit"} complexity, as they involve mathematical calculations and the consideration of multiple relations for a single fund. 
Example questions include: 
\begin{notalistitem}
Q1: What is the ratio of the gross commission against fund net assets for the SFT International Bond Fund?
\end{notalistitem}
\begin{notalistitem}
Q2: What is the ratio of the total purchase sale against fund net assets for the Core Fixed Income Portfolio?
\end{notalistitem}

% Intermediate Example 1: What is the ratio of the {gross commission} against {fund net assets} for fund {X}?\\
% Intermediate Question 1: What is the ratio of the gross commission against fund net assets for the PROFUND VP INTERNET?\\
% Intermediate Answer 1: 0.0001\\

% Intermediate Example 2:What is the ratio of the {total purchase sale} against {fund net assets} for fund {X}?\\
% Intermediate Question 2: What is the ratio of the total purchase sale against fund net assets for the SIIT CORE FIXED INCOME FUND?\\
% Intermediate Answer 2: 7.61 \\   

Similarly to NCEN-QA-Easy, we sampled 200 funds and derived the answers manually from N-CEN reports. We evenly spread the 200 question-answer pairs over the Q1- and Q2-type of questions.

% we created questions by sampling the names of 100 different funds. For each fund, we looked at three different relations, namely (1) total purchase sale, (2) gross commission, and (3) fund average net assets. For instance for a question inquiring about the ratio of total purchase sale against fund net assets, we manually extracted the necessary numerical values and calculated the answers through simple division. This resulted in a total of 200 questions, with 100 variations per the aforementioned examples.

\subsubsection{\textbf{NCEN-QA-Hard}}

In this set, each question focuses on multiple funds, thus presenting a bigger challenge than NCEN-QA-Easy and NCEN-QA-Intermediate. Some example questions are shown below. Q1-type questions require aggregation across multiple funds. Q2-type questions require a full investigation of \textbf{all} reported funds and gather all the funds which are using the specified services (e.g. investment advisor being AlphaMark Advisors), the answer could range from one to multiple funds.

% questions focus on inquiring about a single relation over multiple funds, necessitating the investigation of multiple or all N-CEN reports. The answers to these questions can vary from a single entity name to multiple names. These questions present the greatest level of complexity, as they require consideration of multiple funds and potentially extracting multiple entity names as answers. Examples include:
\begin{notalistitem}
Q1:  What is the gross commission aggregated over funds ClearBridge Dividend Strategy Fund, Baird Mid Cap Growth Fund, and Baron Discovery Fund?
\end{notalistitem}
\begin{notalistitem}
Q2: What funds do the investment advisor company AlphaMark Advisors, LLC manage?
\end{notalistitem}

% \begin{center}
% Hard Example 1: What funds do the {investment advisor} company {X} manage?	\\
% Hard Question 1: What funds do the investment advisor company federated hermes (uk) llp manage?\\
% Intermediate Answer 1: ['FEDERATED HERMES MANAGED VOLATILITY FUND II', 'FEDERATED HERMES GLOBAL TOTAL RETURN BOND FUND']\\

% Hard Example 2: What is the gross commission for funds {X}, {Y}, {Z}?\\
% Hard Question 2: What is the gross commission for funds MASSMUTUAL RETIRESMART BY JPMORGAN 2035 FUND, LORD ABBETT INTERNATIONAL GROWTH FUND, and NEBRASKA TAX-FREE INCOME FUND?\\
% Hard Answer 2: 3280.33\\
% \end{center}

For Q1-type questions, we sampled around 70 tuples of funds and varied the questions between "gross commission" and "total purchase sale". For Q2-type questions, we sampled around 130 service companies across custodians, investment advisors, collateral managers, administrators, and pricing services, and asked about the funds that are using each of these service companies. The longest answer for Q2-type questions in our dataset contained 12 funds. In total, we again had 200 question-answer pairs in this set, same as the other two sets.

% We initially sampled 40 entity names and formulated questions for four specific relations within the N-CEN reports: (1) custodian, (2) investment advisor, (3) administrators, and (4) pricing services. This process led to the creation of 160 questions, following example 1 illustrated above. This is the most complex type of question. since it requires investigating all N-CEN reports. Furthermore, in the second type of hard questions, we randomly selected three fund names and repeated this process for 40 different variations. These questions inquired about the sum of total purchase sales or gross commission for the combination of three funds. Ascertaining the answers to these questions was more complex, as it required investigating three different N-CEN reports and calculating the sum of values for each question. Consequently, we generated an additional 80 questions for this subset.

\subsection{N-CEN APIs}\label{sec:apis}
In FlowMind, we ground the ability of LLMs to reason with reliable Application
Programming Interfaces (APIs), as discussed in Section~\ref{sec:method}. The strength of APIs lies in their robustness, having been designed by domain experts capable of handling vast amounts of data in a structured, parallelized, and deterministic manner. Combining LLMs with APIs stands in stark contrast to solely relying on LLMs alone, which can exhibit limitations such as hallucination and token restrictions.

In our experiments, we developed domain-specific APIs for processing N-CEN reports. FlowMind, in turn, is provided with high-level natural language descriptions of these APIs and is tasked with autonomously generating workflows. These workflows consist of a series of calls to the given APIs, ultimately formulating the answer to the user's query step by step.

We incorporated $6$ crucial APIs into our framework as enumerated in Figure~\ref{fig:recipe}. These APIs encapsulate specific subject matter expertise concerning fund structures and the SEC reporting mechanism, enabling accurate parsing and extraction of information from structured content. Our APIs cover 3 main aspects: \textit{retrieval}, \textit{partition}, and \textit{extract}, as explained below in detail.

\subsubsection{Retrieval}
In order to fetch the correct N-CEN report, we rely on \textit{rapidfuzz}~\cite{rapidfuzz} to disambiguate the input fund name into an N-CEN Accession Number, a unique identifier linking that fund to its most recent report in the \texttt{get\_report} function. \texttt{get\_all\_reports} gathers all the cached results of \texttt{get\_report} applied to every fund name in our dataset. In addition, since a single report may contain multiple funds, \texttt{fetch\_block} can retrieve a single fund block of text within the input report given a specified fund name.

\subsubsection{Partition}
\texttt{segment\_report} function segments an input N-CEN report into a list of fund blocks. This is useful for cases where all funds in a report must be examined for a query.
% To facilitate operators that must iterate through any fund in a particular N-CEN filing, we provide the \texttt{segment\_report} that splits a filing into its component funds, as a list.

\subsubsection{Extract}\label{sec:extract}
\texttt{extract\_entity} and \texttt{extract\_value} are responsible for extracting entity names and certain numbers, respectively. Entities are reported in XML-like blocks with names, identifiers, and other pertinent information such as state, country, classification, etc. Therefore, \texttt{extract\_entity} discovers all named entities in a fund block. We again use \textit{rapidfuzz} to match a query entity label to the ones discovered in a fund block (e.g., `custodian', `collateral manager'). Similarly, \texttt{extract\_value} extracts named values such as `gross commission' and `purchase sales'.
% The \texttt{extract\_value} API searches for the best matching key and returns the value as reported, and is used for items such as commissions, purchase sales, or net assets.
%The need to split this functionality into two functions stems from the structure of the N-CEN data. In order to discover the relevant key to search through, we again employ \textit{rapidfuzz} to find the best matching field to the original query. 

\begin{table*}[t!] 
\centering
\resizebox{\textwidth}{!}{%
\begin{tabular}{>{\centering\arraybackslash} c| >{\centering\arraybackslash}c |>{\centering\arraybackslash}c >{\centering\arraybackslash}c>{\centering\arraybackslash}c|>{\centering\arraybackslash}c>{\centering\arraybackslash}c} 
\hline
& \multicolumn{1}{c|}{\textbf{Baseline}} & \multicolumn{3}{c|}{\textbf{Ablation}} & \multicolumn{2}{c}{\textbf{Proposed}}  \\ \hline
\textbf{Accuracy $\uparrow$} & GPT-Context-Retrieval & FlowMind-NCT & FlowMind-BA & FlowMind-NCP & FlowMind & FlowMind+feedback
\begin{tabular}[c]{@{}c@{}}\end{tabular} \\ \hline
NCEN-QA-Easy & 63.5\% & 88.0\% & 58.0\% & 2.5\% & \textbf{99.5\%} & \textbf{100.0\%}   \\ 
NCEN-QA-Inter & 28.0\% & 91.0\% & 52.5\% & 0.0\% & \textbf{99.0\%} & \textbf{100.0\%}  \\  
NCEN-QA-Hard & 8.5\% & 67.8\% & 28.6\% & 6.5\% & \textbf{89.5\%} & \textbf{96.0\%}   \\ 
\hline
\end{tabular}
}
\caption{Accuracy of outputs from all benchmarked methods. The proposed methods outperformed the baseline significantly. The ablation study reveals the importance of each component in the proposed generic lecture recipe.}
\label{tab:results}
 \vspace{-0.5cm}
\end{table*}

\begin{figure*}[btp]
  \centering
  % \includegraphics[width=0.9\textwidth, height=0.35\textwidth]{example-image-c}
  % Answer: [trim={left bottom right top},clip]
  \includegraphics[clip, trim=0cm 4.65cm 0cm 4cm, width=1.0\textwidth]{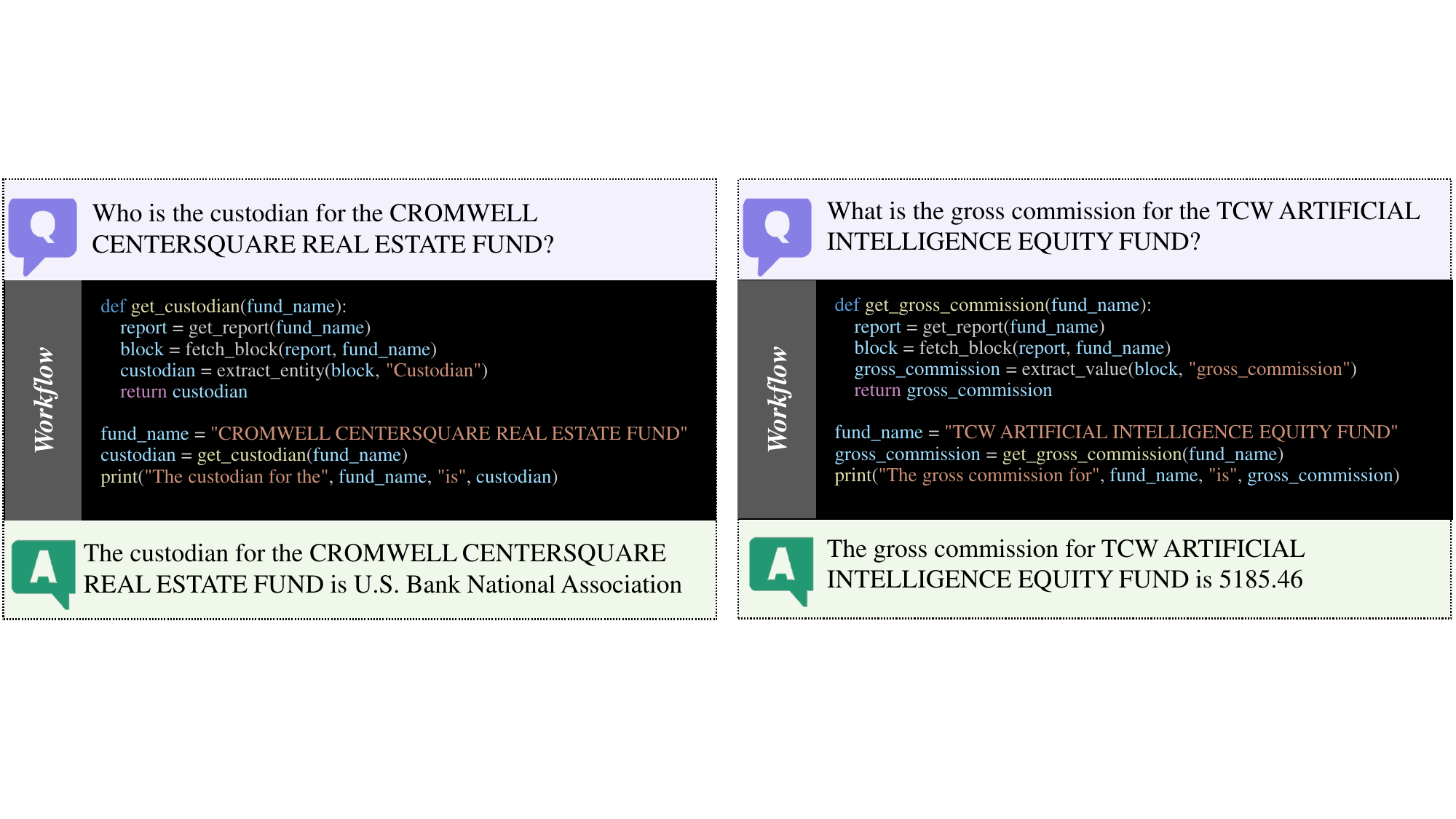}
  \caption{NCEN-QA-Easy: example questions, corresponding workflow and result generated by FlowMind.}
  \label{fig:easy}
\end{figure*}

\subsection{Benchmarks}\label{sec:benchmarks}
% Explain benchmarked methods, potential variants of our method for ablation study
We benchmarked the proposed method FlowMind with and without user feedback, against the commonly adopted methodology of LLMs question answering based on context retrieval (\textbf{GPT-Context-Retrieval}). We also carried out an ablation study where we altered each component in the proposed generic lecture recipe (Section~\ref{sec:recipe}) and benchmarked the performance of each ablation variant of FlowMind. 

\subsubsection{GPT-Context-Retrieval}
A common strategy of question answering without re-training LLMs is to retrieve relevant context to a user question, and prepend the retrieved context to the user question when prompting LLMs~\cite{rubin2021learning, ram2023context, pereira2023visconde}. In our experiments, we used GPT as the LLM and \texttt{text-embedding-ada-002}~\cite{WinNT} to measure the relevancy between a given context and a question. %For simplicity, we will refer to \texttt{text-embedding-ada-002} as \texttt{ada-embed} for the rest of the paper.
First, we used \texttt{get\_all\_reports} and \texttt{segment\_report} (as introduced in Section~\ref{sec:apis}) to get the full list of segmented fund blocks, where each fund block contains the information about a fund. Then, we embed each fund block and the user question into a latent vector using \texttt{text-embedding-ada-002}. We then compare the embedded fund blocks and the embedded user question via cosine similarity to identify the top $k$ fund blocks that are most relevant to the question. These fund blocks are then prepended to the question when prompting GPT to output the answer. We used $k=1$ when evaluating on NCEN-QA-Easy and NCEN-QA-Intermediate because each question is regarding a single fund, and $k=3$ when evaluating on NCEN-QA-Hard because the questions are regarding multiple funds.

Note that some fund blocks are too long during embedding and must be truncated to 8,191 tokens due to the input limit of \texttt{text-embedding-ada-002}. When prompting GPT, we also had to truncate the retrieved top $k$ fund blocks so the prompt is within the 4,096 token limit. These truncations inevitably threw away information that could be useful for question answering.

% First, we embed each individual fund's information from the N-CEN reports, which we refer to as fund blocks using text-embedding-ada-002 \cite{WinNT}. The unique 1536-dimensional embedding vectors represent the semantic meaning of each fund block. To identify the most relevant document sections for a given question, the query is embedded in the same vector space as the fund blocks. Then the cosine similarity or dot product is used to measure the similarity between vectors. Next, the fund blocks are sorted in descending order of relevance and we retrieve the top $k$ most relevant blocks. 

% The next step involves appending the most relevant document sections to the query prompt, with the optimal number of sections determined through experiments. The prompt length is kept as short as possible while providing enough context for the GPT model to generate accurate responses. Finally, we ingest the top $k$ fund blocks along with the user question to retrieve the answer using gpt-3.5-turbo. The GPT model, trained on a vast dataset of natural language processing tasks, uses the provided context to generate contextually relevant answers. The OpenAI Completions API is utilized for this purpose with a low temperature value to ensure more predictable and factual responses. This experiment was applied to the simple, intermediate, and complex questions discussed in prior sections.

\subsubsection{FlowMind-NCT}
The first ablation variant of FlowMind is FlowMind-\textbf{N}o\textbf{C}on\textbf{T}ext, which means we don't provide the context when giving the lecture to GPT in FlowMind. This is to study the effect of the \textit{Context} component as explained in Section~\ref{sec:recipe}. Specifically, we removed the first context sentence from the lecture prompt to GPT.

\subsubsection{FlowMind-BA}
The second ablation variant of FlowMind is FlowMind-\textbf{B}ad\textbf{A}pis, which means we don't provide semantically meaningful names for the input arguments in the APIs' high-level descriptions. This is to study the effect of the \textit{APIs} component as explained in Section~\ref{sec:recipe}. For example, the description of \texttt{get\_report(fund\_name)} becomes \texttt{get\_report(x)}, followed by "Returns the N-CEN report that includes the fund x".

\subsubsection{FlowMind-NCP}
The third ablation variant of FlowMind is FlowMind-\textbf{N}o\textbf{C}ode\textbf{P}rompt, which means we don't explicitly ask GPT to write code. This is to study the effect of the \textit{Code} component as explained in Section~\ref{sec:recipe}. In particular, the last sentence in the lecture prompt to GPT becomes "Wait for user queries, then try to use these functions to respond assuming you have access to these functions. Let me know once you are ready for user queries".

% \subsection{Metric}
% \zz{Discuss how we measure the accuracy of the derived answer. Three types of answers: a name string, multiple name strings, and a number. For each type, discuss the way we measure the accuracy. string matches (LEI?), number has to be exact same to ground truth to be counted as correct answer}

\begin{figure*}[t!]
  \centering
   % Answer: [trim={left bottom right top},clip]
  \includegraphics[clip, trim=0cm 2.95cm 0cm 4cm, width=1.0\textwidth]{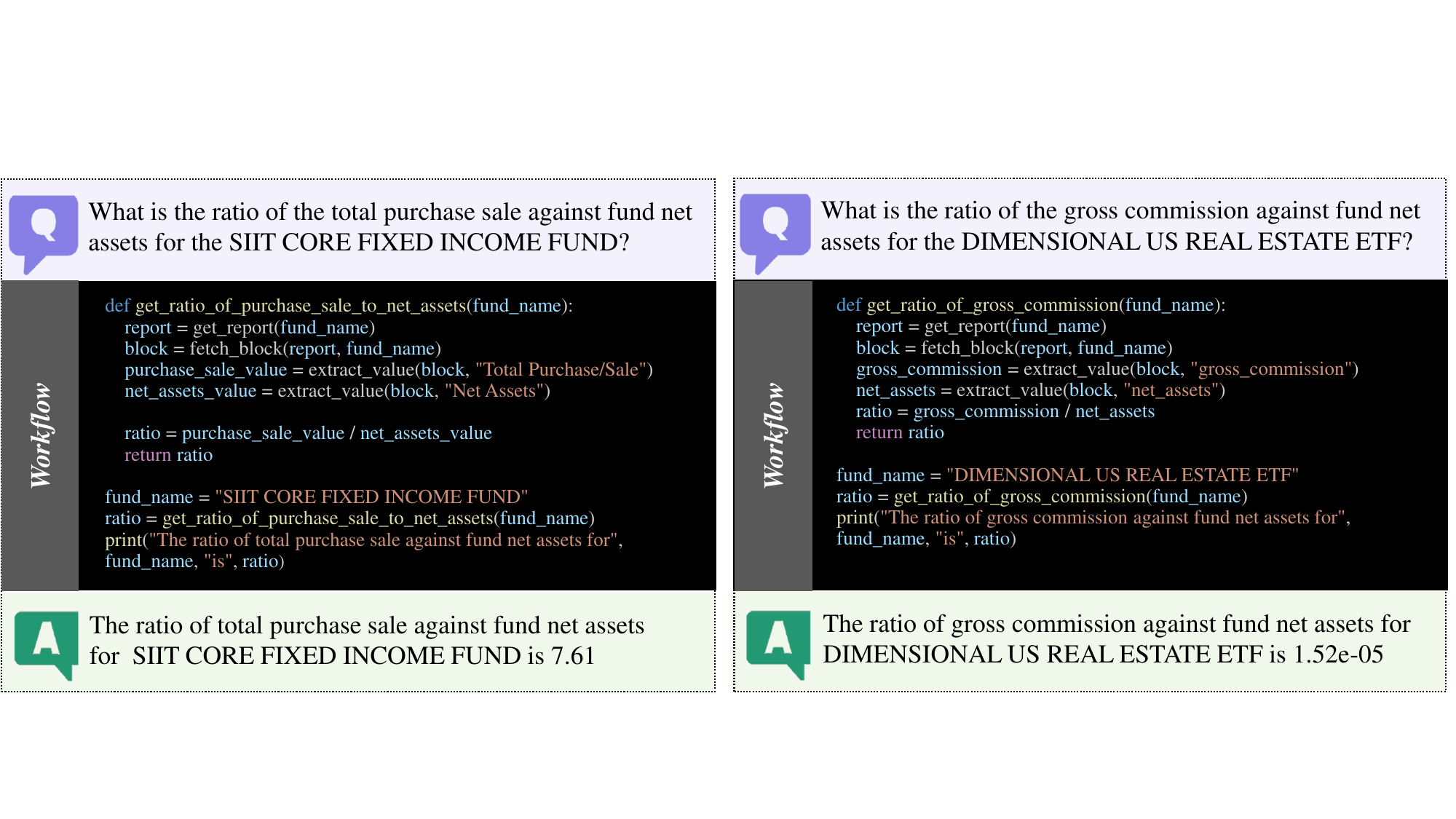}
  \caption{NCEN-QA-Intermediate: example questions, corresponding workflows and results generated by FlowMind.}
  \label{fig:intermediate}
\end{figure*}

\begin{figure*}[t!]
  \centering
   % Answer: [trim={left bottom right top},clip]
  \includegraphics[clip, trim=0cm 0.2cm 0cm 0.7cm, width=1.0\textwidth]{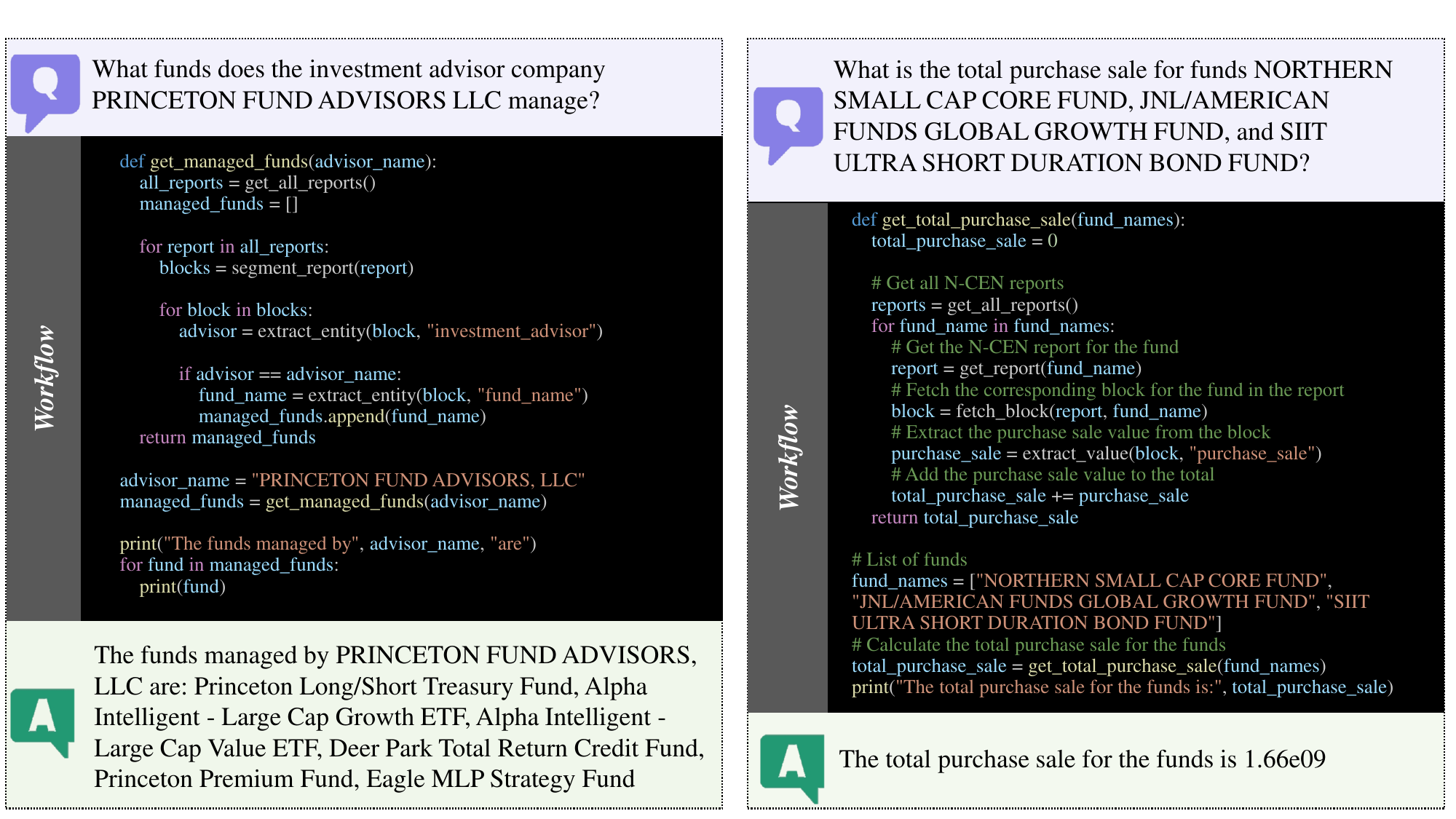}
  \caption{NCEN-QA-Hard: example questions, corresponding workflows and answers generated by FlowMind.}
  \label{fig:hard}
\end{figure*}

\subsection{Results}\label{sec:results}
% discuss results and take home message
We evaluated the proposed methods and all the methods mentioned in Section~\ref{sec:benchmarks} using the curated NCEN-QA dataset. We report the benchmarked accuracy of each method in Table~\ref{tab:results}. The accuracy was measured by comparing the output answer to the ground truth. For questions related to entity names, an answer is considered accurate only if it captures all the entity names in the ground truth. For questions related to numbers, an answer was deemed accurate if it matched the ground truth number after rounding to the ground truth's precision.

Our results revealed that FlowMind, even without user feedback, significantly outperformed the GPT-Context-Retrieval baseline method. We show some qualitative examples of FlowMind in Figure~\ref{fig:easy},~\ref{fig:intermediate},~\ref{fig:hard}. The baseline's performance notably degrades as the complexity of questions increases, often due to 1) truncation on the retrieved context to fit within GPT input token limits, 2) hallucination, 3) retrieval of incorrect context, and 4) inaccurate number calculation.

In our ablation study, we found each component of the lecture recipe to be vital for FlowMind's success. FlowMind-NCT, which excluded context in the lecture prompt, performed worse than the standard FlowMind, emphasizing the need for context. FlowMind-BA's poor performance highlighted the importance of quality API descriptions with semantically meaningful argument names; it often failed due to incorrect API calls or use of wrong arguments. FlowMind-NCP, which did not generate any executable workflow code in most cases, yielded the lowest accuracy, indicating the necessity of explicit prompting LLM to write code to express the workflow that derives the answer.

\begin{figure*}[t!]
  \centering
   % Answer: [trim={left bottom right top},clip]
  \includegraphics[clip, trim=0cm 3.3cm 0cm 1cm, width=1.0\textwidth]{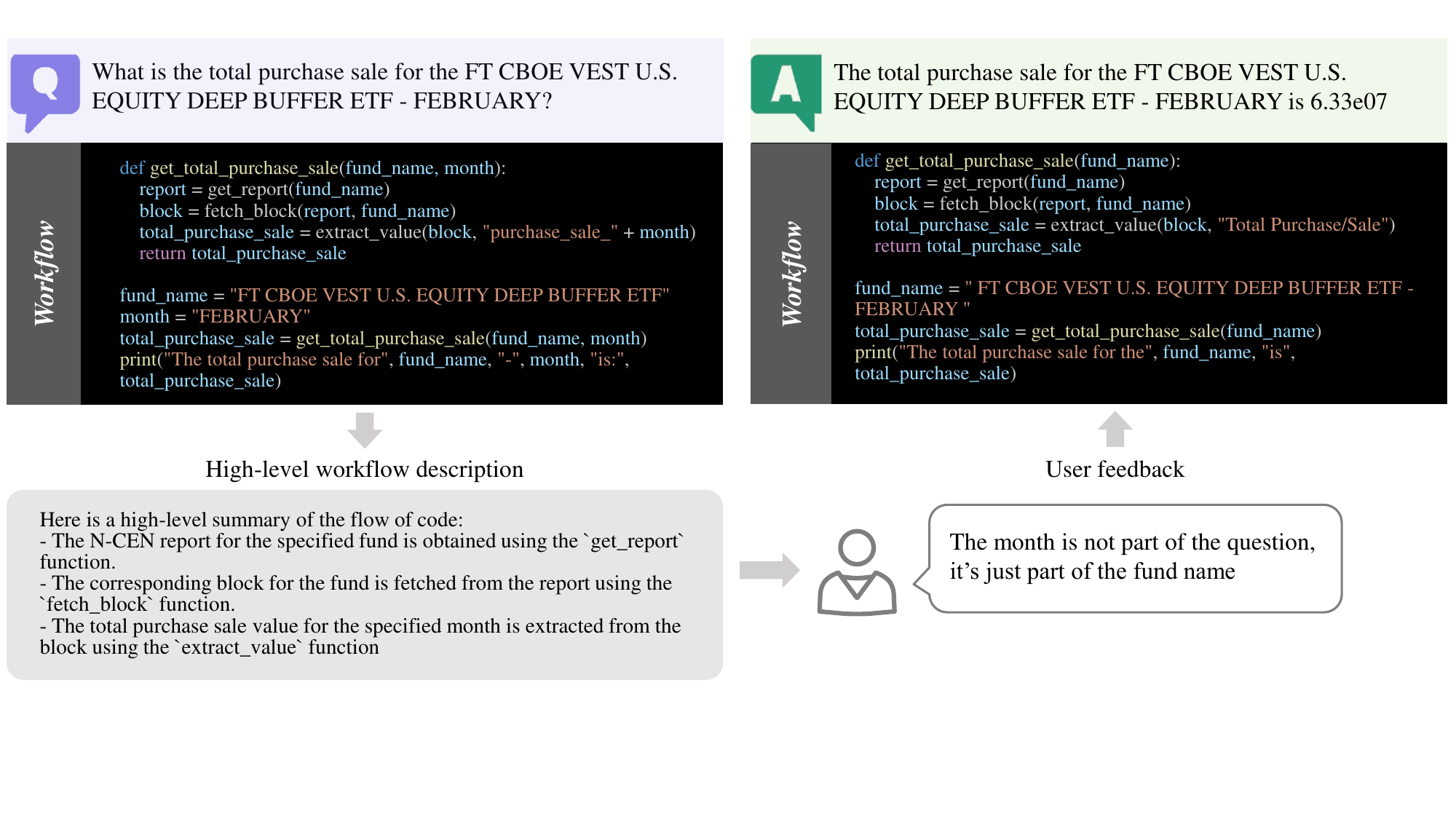}
  \caption{Example of FlowMind correcting workflow given user feedback.}
  \label{fig:feedback}
\end{figure*}

We further improved FlowMind's performance by incorporating user feedback, allowing it to update the workflow after generating the first executable workflow code based on user suggestions. This led to increased accuracy, as evident in NCEN-QA-Easy shown in Figure~\ref{fig:feedback}, where a question was about the purchase sale of a fund with `February' in its name. Initially, FlowMind misunderstood the question, interpreting 'February' as the month of a fund's purchase sale, rather than recognizing it as part of the fund's name. With user feedback, however, this confusion was clarified, and FlowMind adjusted its workflow to correctly interpret 'February' as part of the fund name, not a time reference. For NCEN-QA-Intermediate and NCEN-QA-Hard, user feedback helped correct false assumptions regarding the information that could be extracted using the \textit{extract} APIs (Section ~\ref{sec:extract}). For instance, FlowMind initially operated under the assumption that `gross commission ratio' was a value that could be directly extracted, when it actually required computation. Another example was with the terms `purchase' and `sale', which FlowMind initially considered as separate values for extraction. However, they were part of a single term, "purchase sale", that needed to be extracted as a whole. FlowMind was able to correct its workflow based on user feedback, thus boosting its performance to 100\% or near 100\% across the datasets.

% \ww{Discuss ambiguity in names, like in questions of what funds are managed by company X, that name X can mean many different entities if not specific, provide examples on the Stanbic Bank case. Within our scope of the paper, we assume the questions are addressing the exact name of the entity without ambiguity except for lower versus capital case difference and extra spaces - if we have time, as we use the exact name for synthetic data generation}

\section{Conclusion}
% This user-friendly system also integrates human intervention, allowing users, even those with minimal expertise, to inspect, provide feedback, and refine the generated workflow to suit their unique task requirements.

% +discussion: talks about if APIs are corresponding to mouse clicks and keyboard strokes, that will work well for repetitive tasks such as moving files and so on

In conclusion, our work presents a significant leap in using LLMs for automatic workflow generation. By combining lecture prompt design, user feedback, and secure, grounded reasoning, FlowMind provides a reliable, adaptable, and efficient solution for handling spontaneous tasks with auto-genenated workflows. Our work opens new avenues for more widespread adoption of LLMs, particularly in industries where data security and the spontaneity of tasks are of paramount importance.

We also introduce a new finance dataset NCEN-QA, which serves as a robust benchmark platform for automatic workflow generation systems on N-CEN reports question-answering tasks about funds, thus providing a valuable resource for the broader research community.

In the future, it's worth investigating crowdsourcing user feedback to refine workflows in FlowMind at scale, as well as life-long learning over past user-approved examples to evolve its performance over time. In addition, FlowMind can be expanded in the future to handle big libraries of APIs by retrieving the most relevant APIs for a given task given embedding similarity.

% \zz{future: could crowdsource user feedbacks; learn from past examples; auto correct from execution errors.}
% \zz{talk about future directions on handling a large library and use embedding similarity retrieval to get more relevant APIs.}

\begin{acks}
This paper was prepared for informational purposes by the Artificial Intelligence Research group of JPMorgan Chase \& Co and its affiliates (“J.P. Morgan”) and is not a product of the Research Department of J.P. Morgan.  J.P. Morgan makes no representation and warranty whatsoever and disclaims all liability, for the completeness, accuracy or reliability of the information contained herein.  This document is not intended as investment research or investment advice, or a recommendation, offer or solicitation for the purchase or sale of any security, financial instrument, financial product or service, or to be used in any way for evaluating the merits of participating in any transaction, and shall not constitute a solicitation under any jurisdiction or to any person, if such solicitation under such jurisdiction or to such person would be unlawful.  
\end{acks}
\bibliographystyle{ACM-Reference-Format}
\bibliography{egbib}

\end{document}